\documentclass[11pt,a4paper]{article}

\usepackage[hyperref]{myacl}
\usepackage{times}
\usepackage{latexsym}

\usepackage[utf8]{inputenc} 
\usepackage[T1]{fontenc}    
\usepackage[english]{babel}
\usepackage{amssymb, amsmath}
\usepackage{url}            
\usepackage{relsize}        
\usepackage[skip=6pt]{caption}
\usepackage{booktabs}       
\usepackage{amsfonts}       
\usepackage{nicefrac}       
\usepackage{microtype}      
\usepackage[pdftex]{graphicx}
\usepackage{hyperref}       

\aclfinalcopy 

\title{Context encoders as a simple but powerful extension of word2vec}

\author{Franziska Horn\\
  Machine Learning Group\\
  Technische Universität Berlin, Germany \\
  \texttt{franziska.horn@campus.tu-berlin.de}\\
}

\begin{document}
\setlength{\abovedisplayskip}{0pt}
\setlength{\belowdisplayskip}{4pt}
\setlength{\intextsep}{8pt plus 3pt minus 2pt}
\setlength{\textfloatsep}{9pt plus 3pt minus 2pt}
\setlength{\floatsep}{5pt plus 3pt minus 2pt}
\setlength{\dblfloatsep}{7pt plus 3pt minus 2pt}
\setlength{\dbltextfloatsep}{11pt plus 3pt minus 2pt}

\maketitle

\begin{abstract}
With a simple architecture and the ability to learn meaningful word embeddings efficiently from texts containing billions of words, word2vec remains one of the most popular neural language models used today. However, as only a single embedding is learned for every word in the vocabulary, the model fails to optimally represent words with multiple meanings. Additionally, it is not possible to create embeddings for new (out-of-vocabulary) words on the spot. Based on an intuitive interpretation of the continuous bag-of-words (CBOW) word2vec model's negative sampling training objective in terms of predicting context based similarities, we motivate an extension of the model we call context encoders (ConEc). By multiplying the matrix of trained word2vec embeddings with a word's average context vector, out-of-vocabulary (OOV) embeddings and representations for a word with multiple meanings can be created based on the word's local contexts. The benefits of this approach are illustrated by using these word embeddings as features in the CoNLL 2003 named entity recognition (NER) task.
\end{abstract}

\section{Introduction}
Representation learning is very prominent in the field of natural language processing (NLP). For example, word embeddings learned by neural language models (NLM) were shown to improve the performance when used as features for supervised learning tasks such as named entity recognition (NER) \citep{collobert2011natural,turian2010word}. The popular \emph{word2vec} model \citep{mikolov2013efficient,mikolov2013distributed} learns meaningful word embeddings by considering only the words' local contexts. Thanks to its shallow architecture it can be trained very efficiently on large corpora. The model, however, only learns a single representation for words from a fixed vocabulary. Consequently, if in a task we encounter a new word that was not present in the texts used for training, we cannot create an embedding for this word without repeating the time consuming training procedure of the model.\footnote{In practice the model is trained on such a large vocabulary that it is rare to encounter a word that does not have an embedding. Yet there are still scenarios where this is the case, for example, it is unlikely that the term ``W10281545'' is encountered in a regular training corpus, but we might still want its embedding to represent a search query like ``whirlpool W10281545 ice maker part''.} Furthermore, a single embedding does not optimally represent a word with multiple meanings. For example, ``Washington'' is both the name of a US state as well as a former president and only by taking into account the word's local context can one identify the proper sense.

Based on an intuitive interpretation of the continuous bag-of-words (CBOW) word2vec model's negative sampling training objective, we propose an extension of the model we call \emph{context encoders} (ConEc). This allows for an easy creation of OOV embeddings as well as a better representation of words with multiple meanings by simply multiplying the trained word2vec embeddings with the words' average context vectors. As demonstrated by the CoNLL 2003 NER challenge, the classification performance can be significantly improved when using as features the word embeddings created with ConEc instead of word2vec.

\paragraph{Related work} In the past, NLM have addressed the issue of polysemy in various ways. For example, sense2vec is an extension of word2vec, where in a preprocessing step all words in the training corpus are annotated with their part-of-speech (POS) tag and then the embeddings are learned for tokens consisting of the words themselves and their POS tags. This way, different representations are generated e.g.~for words that are used both as a noun and verb \citep{trask2015sense2vec}. Other methods first cluster the contexts in which the words appear \citep{huang2012improving} or use additional resources such as wordnet to identify multiple meanings of words \citep{rothe2015autoextend}. One possibility to create OOV embeddings is to learn representations for all character n-grams in the texts and then compute the embedding of a word by combining the embeddings of the n-grams occurring in it \citep{bojanowski2016enriching}. However, none of these NLM are designed to solve both the OOV and polysemy problem at the same time. Furthermore, compared to word2vec they require more parameters, resources, or additional steps in the training procedure. ConEc on the other hand can generate OOV embeddings as well as improved representations for words with multiple meanings by simply multiplying the matrix of trained word2vec embeddings with the words' average context vectors.

\section{Background: CBOW word2vec trained with negative sampling}
Word2vec (Fig.~\ref{fig:word2vec} in the Appendix) learns $d\,$-dimensional vector representations, referred to as word embeddings, for all $N$ words in the vocabulary. It is a shallow NLM with parameter matrices $W_0, W_1 \in \mathbb{R}^{N\times d}$, which are tuned iteratively by scanning huge amounts of text sentence by sentence. Based on some context words, the algorithm tries to predict the target word between them. Mathematically, this is realized by first computing the sum of the embeddings of the context words by selecting the appropriate rows from $W_0$. This vector is then multiplied by several rows selected from $W_1$: one of these rows corresponds to the target word, while the others correspond to $k$ `noise' words selected at random (negative sampling). After applying a non-linear activation function, the backpropagation error is computed by comparing this output to a label vector $\mathbf{t} \in \mathbb{R}^{k+1}$, which is 1 at the position of the target word and 0 for all $k$ noise words. After the training of the model is complete, the word embedding for a target word is the corresponding row of $W_0$.

\section{Context Encoders}
Similar words appear in similar contexts \citep{harris1954distributional}. For example, two words synonymous with each other could be exchanged for one another in almost all contexts without a reader noticing. Based on the context word co-occurrences, pairwise similarities between all $N$ words of the vocabulary can be computed, resulting in a similarity matrix $S \in \mathbb{R}^{N\times N}$ (or for a single word $w$ the vector $\mathbf{s}_w \in \mathbb{R}^N$) with similarity scores between $0$ and $1$. 
These similarities should be preserved in the word embeddings, e.g.~the cosine similarity between the embedding vectors of two words used in similar contexts should be close to $1$, or, more generally, the scalar product of the matrix with word embeddings $Y \in \mathbb{R}^{N\times d}$ should approximate $S$. Obviously, the most straightforward way of obtaining word embeddings satisfying $YY^\top\approx S$ would be to compute the singular value decomposition (SVD) of the similarity matrix $S$ and use the eigenvectors corresponding to the $d$ largest eigenvalues \cite{levy2014linguistic,levy2015improving}. As our vocabulary typically comprises tens of thousands of words, performing an SVD of the corresponding similarity matrix is computationally far too expensive. Yet, while the similarity matrix would be huge, it would also be quite sparse, as many words are of course not synonymous with each other. If we picked a small number $k$ of random words, chances are their similarities to a target word would be close to $0$. Therefore, while the product of a single word's embedding $\mathbf{y}_w \in \mathbb{R}^d$ and the matrix of all embeddings $Y$ should result in a vector $\mathbf{\hat s}_w \in \mathbb{R}^N$ close to the true similarities $\mathbf{s}_w$ of this word, if we only consider a small subset of $\mathbf{\hat s}_w$ corresponding to the word itself and $k$ random words, it is sufficient if this approximates the binary vector $\mathbf{t}_w \in \mathbb{R}^{k+1}$, which is $1$ for the word itself and $0$ elsewhere.

\begin{figure}[h!]
  \centering
    \includegraphics[width=\linewidth]{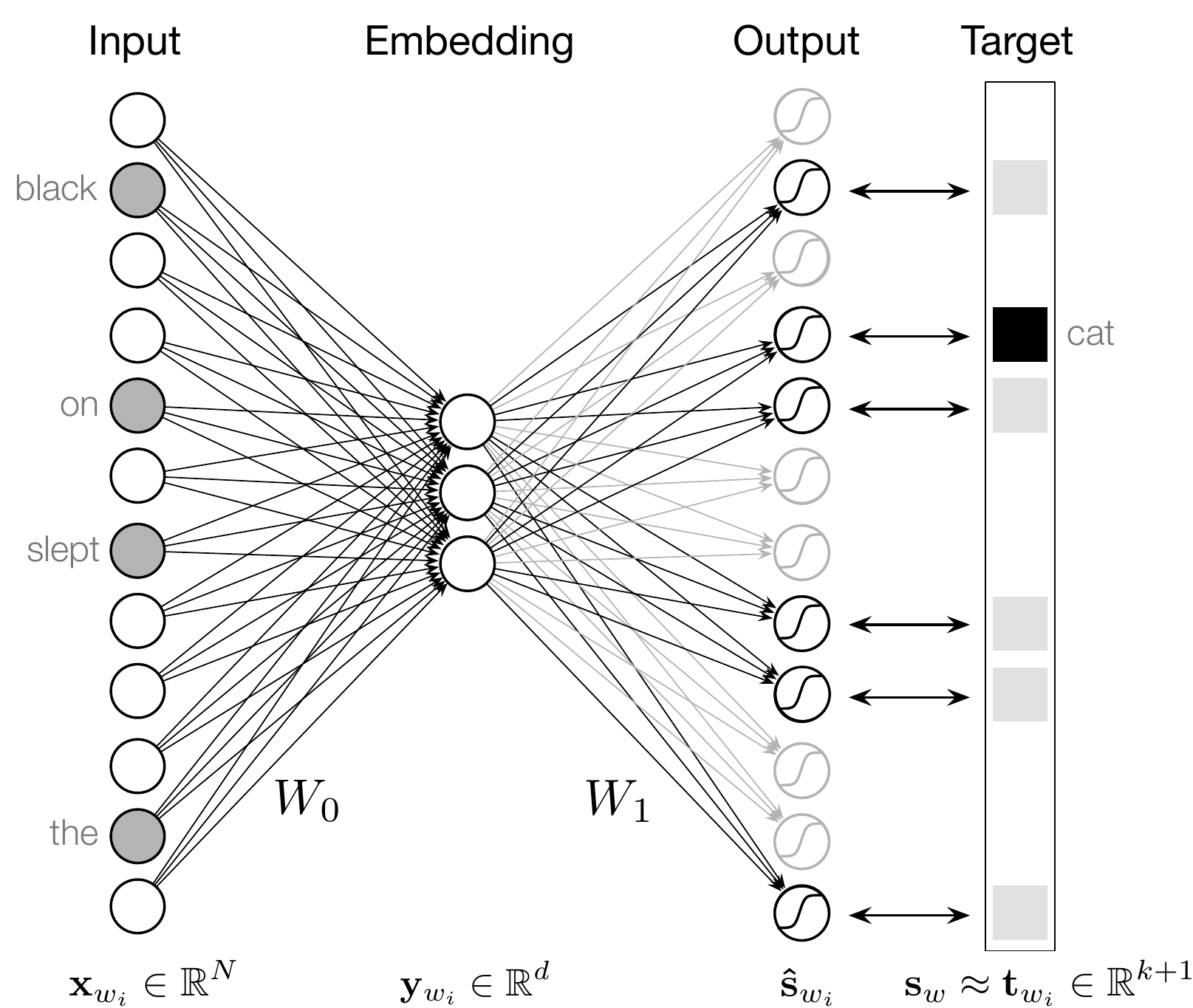}
  \caption{Context encoder (ConEc) NN architecture corresponding to the CBOW word2vec model trained with negative sampling. 
  }
  \label{fig:contextenc}
\end{figure}

The CBOW word2vec model trained with negative sampling can therefore be interpreted as a neural network (NN) that predicts a word's similarities to other words (Fig.~\ref{fig:contextenc}). During training, for each occurrence $i$ of a word $w$ in the texts, a binary vector $\mathbf{x}_{w_i}\in\mathbb{R}^N$, which is $1$ at the positions of the context words of $w$ and $0$ elsewhere, is used as input to the network and multiplied by a set of weights $W_0$ to arrive at an embedding $\mathbf{y}_{w_i} \in \mathbb{R}^d$ (the summed rows of $W_0$ corresponding to the context words). This embedding is then multiplied by another set of weights $W_1$, which corresponds to the full matrix of word embeddings $Y$, to produce the output of the network, a vector $\mathbf{\hat s}_{w_i} \in \mathbb{R}^N$ containing the approximated similarities of the word $w$ to all other words. The training error is then computed by comparing a subset of the output to a binary target vector $\mathbf{t}_{w_i} \in \mathbb{R}^{k+1}$, which serves as an approximation of the true similarities $\mathbf{s}_w$ when considering only a small number of random words. We refer to this interpretation of the model as \emph{context encoders} (ConEc), as it is closely related to similarity encoders (SimEc), a dimensionality reduction method used for learning similarity preserving representations of data points \citep{horn2017simec}. 

While the training procedure of ConEc is identical to that of word2vec, there is a difference in the computation of a word's embedding after the training is complete. In the case of word2vec, the word embedding is simply the row of the tuned $W_0$ matrix. When considering the idea behind the optimization procedure, we instead propose to create the representation of a target word $w$ by multiplying $W_0$ with the word's average context vector $\mathbf{x}_{w}$, as this better resembles how the word embeddings are computed during training.

We distinguish between a word's `global' and `local' average context vector (CV): The global CV is computed as the average of all binary CVs $\mathbf{x}_{w_i}$ corresponding to the $M_w$ occurrences of $w$ in the whole training corpus:
\begin{align*}
\mathbf{x}_{w_\text{global}} = {1\over M_w} \sum_{i=1}^{M_w} \mathbf{x}_{w_i},
\end{align*}
while the local CV $\mathbf{x}_{w_\text{local}}$ is computed likewise but considering only the $m_w$ occurrences of $w$ in a single document.
We can now compute the embedding of a word $w$ by multiplying $W_0$ with the weighted average between both CVs:
\begin{align} \label{eq:mix}
 \mathbf{y}_w = (a\cdot\mathbf{x}_{w_\text{global}} + (1-a)\,\mathbf{x}_{w_\text{local}})^\top W_0
\end{align}
with $a \in [0,1]$. The choice of $a$ determines how much emphasis is placed on the word's local context, which helps to distinguish between multiple meanings of the word \citep{melamud2015modeling}.\footnote{This implicitly assumes a word is only used in a single sense in one document.} As an out-of-vocabulary word does not have a global CV (as it never occurred in the training corpus), its embedding is computed solely based on the local context, i.e. setting $a=0$.

With this new perspective on the model and optimization procedure, another advancement is feasible. Since the context words are merely a sparse feature vector used as input to a NN, there is no reason why this input vector should not contain other features about the target word as well. For example, the feature vector $\mathbf{x}_w$ could be extended to contain information about the word's case, part-of-speech (POS) tag, or other relevant details. While this would increase the dimensionality of the first weight matrix $W_0$ to include the additional features when mapping the input to the word's embedding, the training objective and therefore also $W_1$ would remain unchanged. These additional features could be especially helpful if details about the words would otherwise get lost in preprocessing (e.g.~by lowercasing) or to retain information about a word's position in the sentence, which is ignored in a BOW approach. These extended ConEcs are expected to create embeddings that even better distinguish between the words' different senses by taking into account, for example, if the word is used as a noun or verb in the current context, similar to the sense2vec algorithm \citep{trask2015sense2vec}. But instead of explicitly learning multiple embeddings per term, like sense2vec, only the dimensionality of the input vector is increased to include the POS tag of the current word as a feature, which is expected to improve generalization if few training examples are available.

\section{Experiments}
\begin{figure*}[!h]
  \centering
      \includegraphics[height=6.2cm]{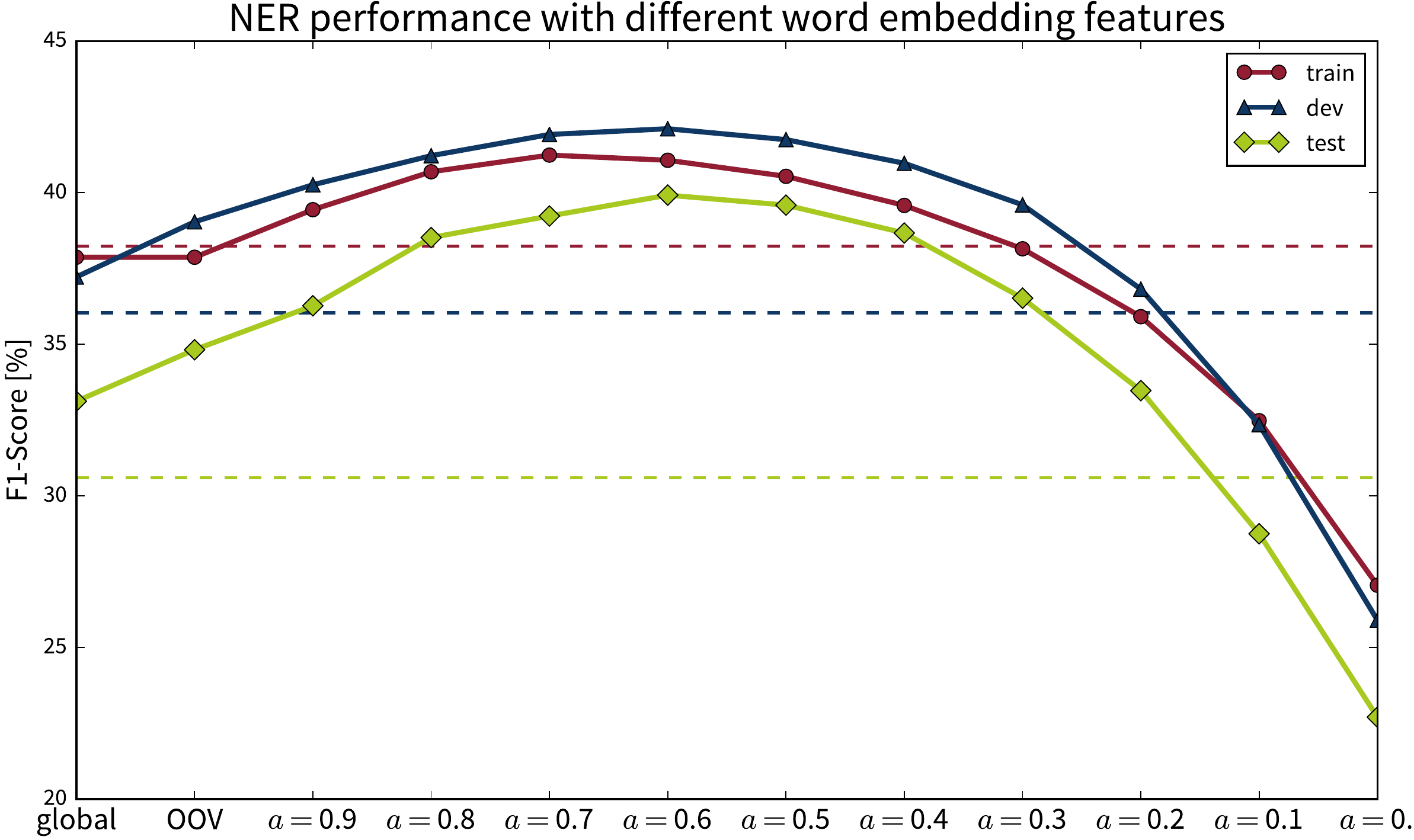}
      \includegraphics[height=6.2cm]{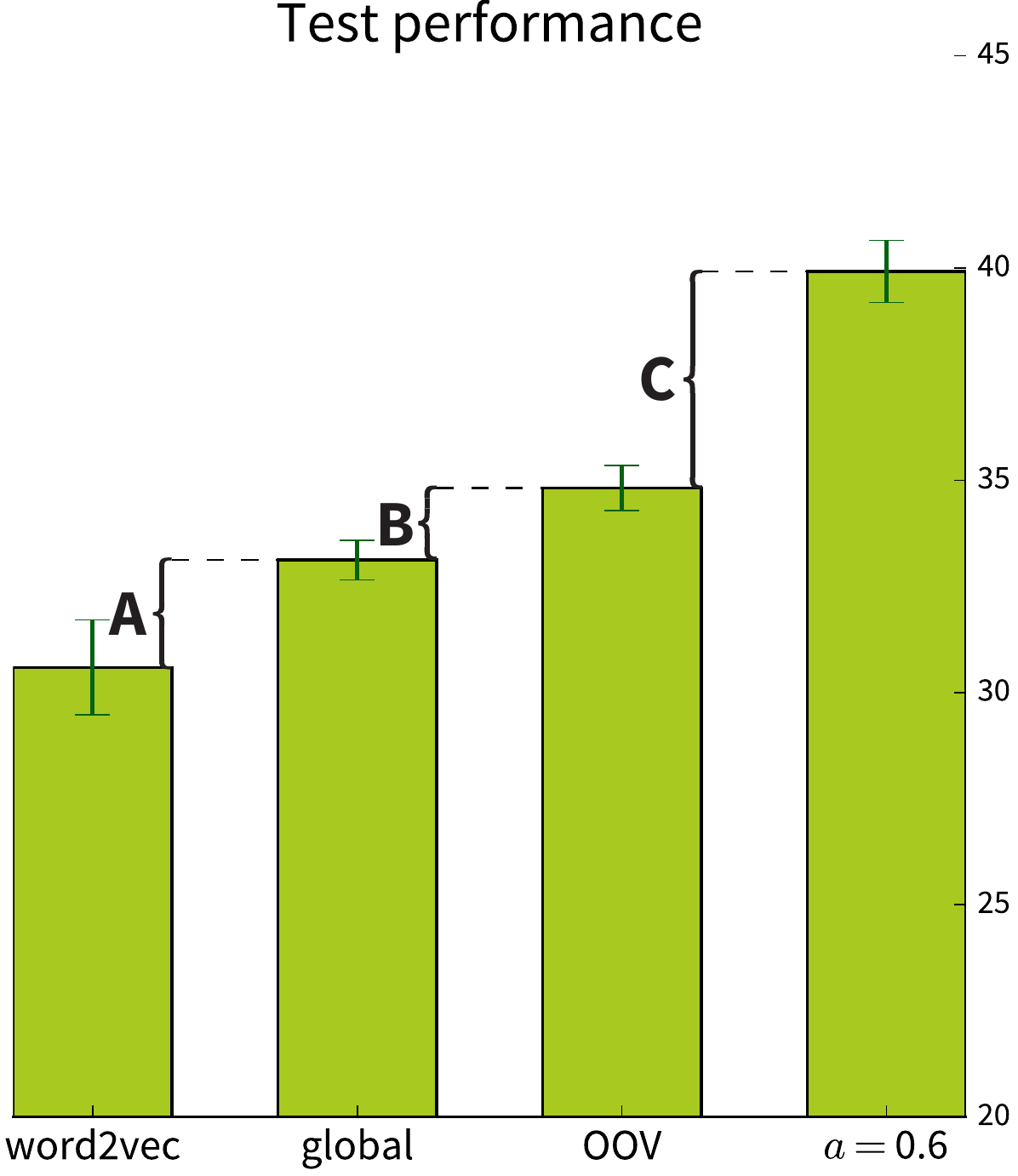}
  \caption{Results of the NER task based on three random initializations of the word2vec model. \emph{Left panel:} Overall results, where the mean performance using word2vec embeddings (\emph{dashed lines}) is considered as our baseline, all other embeddings are computed with ConEcs using various combinations of the words' global and local CVs. \emph{Right panel:} Increased performance (mean and standard deviation) on the test fold when using ConEc: Multiplying the word2vec embeddings with global CVs yields a performance gain of $2.5$ percentage points (\emph{A}). By additionally using local CVs to create OOV word embeddings we gain another $1.7$ points (\emph{B}). When using a combination of global and local CVs (with $a=0.6$) to distinguish between the different meanings of words, the F1-score increases by another $5.1$ points (\emph{C}), yielding a F1-score of $39.92\%$, which marks a significant improvement compared to the $30.59\%$ reached with word2vec features.}
  \label{fig:ner_results}
\end{figure*}
The word embeddings learned by word2vec and context encoders are evaluated on the CoNLL 2003 NER benchmark task \citep{conll2003}. We use a CBOW word2vec model trained with negative sampling as described above where $k=13$, the embedding dimensionality $d$ is $200$ and we use a context window of $5$ words. The word embeddings created by ConEc are built directly on top of the word2vec model by multiplying the original embeddings ($W_0$) with the respective context vectors. Code to replicate the experiments is available online.\footnote{\url{https://github.com/cod3licious/conec}} Additionally, the performance on a word analogy task \citep{mikolov2013efficient} is reported in the Appendix.

\paragraph{Named Entity Recognition}
The main advantage of context encoders is their ability to use local context to create OOV embeddings and distinguish between the different senses of words. The effects of this are most prominent in a task such as NER, where the local context of a word can make all the difference, e.g.~to distinguish between the ``Chicago Bears'' (an organization) and the city of Chicago (a location). We tested this on the CoNLL 2003 NER task by using the word embeddings as features together with a logistic regression classifier. The reported F1-scores were computed using the official evaluation script. The results achieved with various word embeddings in the training, development, and test part of the CoNLL task are reported in Fig.~\ref{fig:ner_results}.
It should be noted that we are using this task as an extrinsic evaluation to illustrate the advantages of ConEc embeddings over the regular word2vec embeddings. To isolate the effects on the performance, we are only using these word embeddings as features, while typically the performance on this NER challenge is much higher when other features such as a word's case or POS tag are included as well.

The word2vec embeddings were trained on the documents used in the training part of the task. OOV words in the development and test parts are represented as zero vectors.\footnote{Since this is a very small corpus, we trained word2vec for 25 iterations on these documents.}  
With three parameter settings, we illustrate the advantages of ConEc:\\
\textsl{A) Multiplying the word2vec embeddings by the words' average context vectors generally improves the embeddings.} To show this, ConEc word embeddings were computed using only global CVs (Eq.~\ref{eq:mix} with $a=1$), which means OOV words again have a zero representation. With these embeddings (labeled `global' in Fig.~\ref{fig:ner_results}), the performance improves on the dev and test folds of the task.\\
\textsl{B) Useful OOV embeddings can be created from the local context of a new word.} To show this, the ConEc embeddings for words from the training vocabulary ($w \in N$) were computed as in \textsl{A)}, but now the embeddings for OOV words ($w' \notin N$) were computed using local CVs (Eq.~\ref{eq:mix} with $a=1 \;\forall\, w \in N$ and $a=0 \;\forall\, w' \notin N$; referred to as `OOV' in the figure). The training performance obviously stays the same, because here all words have an embedding based on their global contexts. However, there is a jump in the ConEc performance on the dev and test folds, where OOV words now have a representation based on their local contexts.\\
\textsl{C) Better embeddings for a word with multiple meanings can be created by using a combination of the word's average global and local CVs as input to the ConEc.} To show this, the OOV embeddings were computed as in \textsl{B)}, but now for the words occurring in the training vocabulary, the local context was taken into account as well by setting $a < 1$ (Eq.~\ref{eq:mix} with $a \in [0,1) \;\forall\, w \in N$ and $a=0 \;\forall\, w' \notin N$). The best performances on all folds are achieved when averaging the global and local CVs with around $a=0.6$ before multiplying them with the word2vec embeddings. This clearly shows that ConEc embeddings created by incorporating local context can help distinguish between multiple meanings of words.

\section{Conclusion}
Context encoders are a simple but powerful extension of the CBOW word2vec model trained with negative sampling. By multiplying the matrix of trained word2vec embeddings with the words' average context vectors, ConEcs are easily able to create OOV embeddings on the spot as well as distinguish between multiple meanings of words based on their local contexts. The benefits of this were demonstrated in the CoNLL NER challenge.

\section*{Acknowledgments}
I would like to thank Antje Relitz, Ivana Balažević, Christoph Hartmann, Andreas Nowag, Klaus-Robert Müller, and other anonymous reviewers for their helpful comments on earlier versions of this manuscript.\\ Franziska Horn acknowledges funding from the Elsa-Neumann scholarship from the TU Berlin.

\bibliographystyle{acl_natbib}
\bibliography{../../phd_collected}

\appendix

\section*{Appendix}
\begin{figure*}[!h]
  \centering
    \includegraphics[width=0.82\linewidth]{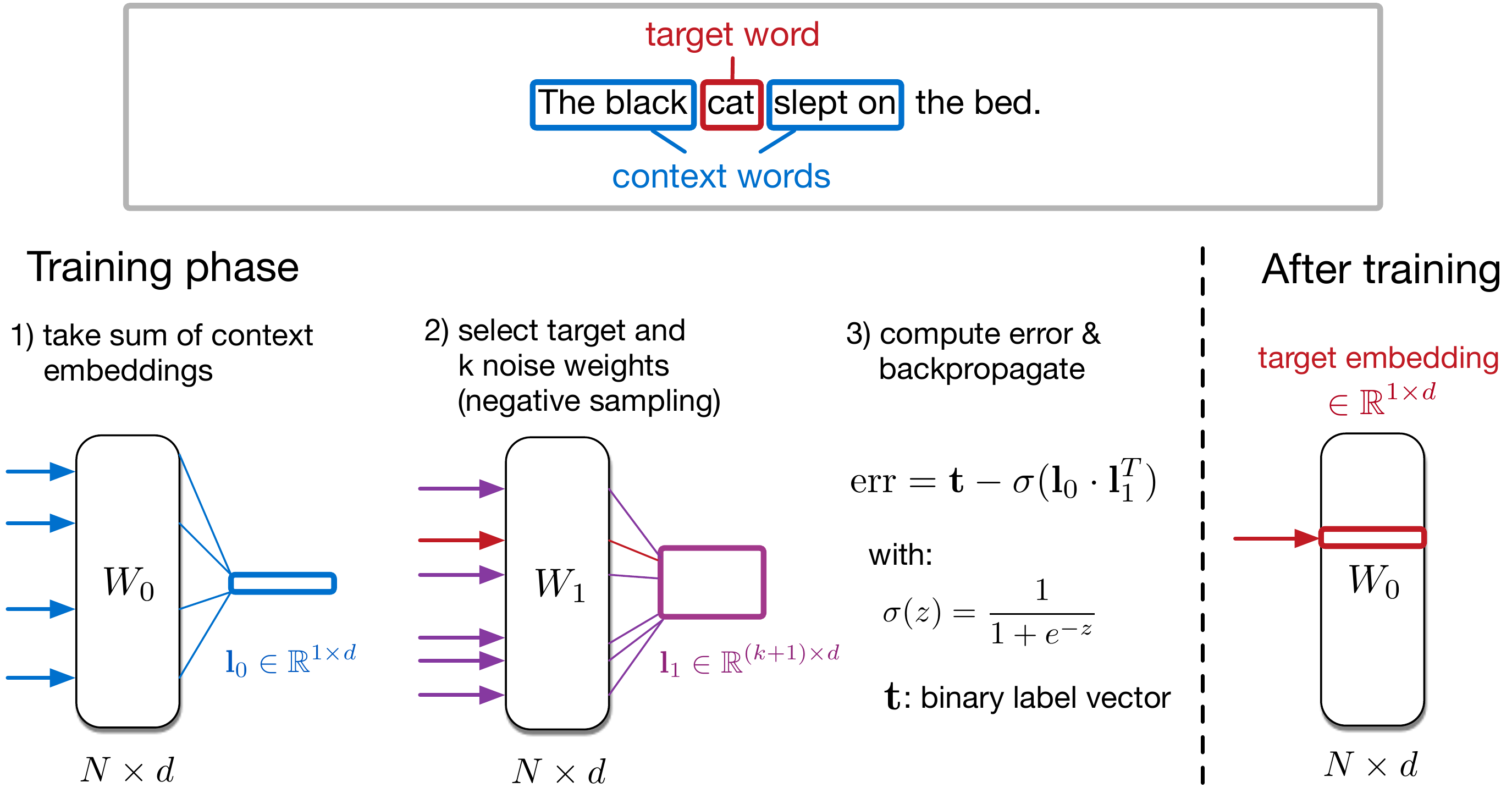}
  \caption{Continuous bag-of-words (CBOW) word2vec model trained with negative sampling \citep{mikolov2013efficient,mikolov2013distributed,goldberg2014word2vec}.}
  \label{fig:word2vec}
\end{figure*}

\paragraph{Analogy task}
To show that the word embeddings created with context encoders capture meaningful semantic and syntactic relationships between words, we evaluated them on the original analogy task published together with the word2vec model \citep{mikolov2013efficient}.\footnote{\url{https://code.google.com/archive/p/word2vec/}} This task consists of many questions in the form of ``\emph{man} is to \emph{king} as \emph{woman} is to XXX'' where the model is supposed to find the correct answer \emph{queen}. This is accomplished by taking the word embedding for \emph{king}, subtracting from it the embedding for \emph{man} and then adding the embedding for \emph{woman}. This new word vector should then be most similar (with respect to the cosine similarity) to the embedding for \emph{queen}.\footnote{Readers familiar with \citet{levy2015improving} will recognize this as the 3CosAdd method. We have tried 3CosMul as well, but found that the results did not improve significantly and therefore omitted them here.}
The word2vec model was trained for ten iterations on the \texttt{text8} corpus,\footnote{\url{http://mattmahoney.net/dc/text8.zip}} which contains around 17 million words and a vocabulary of about 70k unique words, as well as the training part of the \texttt{1-billion} benchmark dataset,\footnote{\url{http://code.google.com/p/1-billion-word-language-modeling-benchmark/}} which contains over 768 million words with a vocabulary of 486k unique words.\footnote{In this experiment we ignore all words that occur less than 5 times in the training corpus.} The ConEc embeddings were then constructed by multiplying the word2vec embeddings with the words' average global context vectors obtained from the same corpus as the word2vec model was trained on. To achieve the best results, we also had to include the target word itself in these context vectors.

The results of the analogy task are shown in Table~\ref{analogy}. To capture some of the semantic relations between words (e.g. the first four task categories) it can be advantageous to use context encoders instead of word2vec. One reason for the ConEcs' superior performance on some of the task categories, but not others, might be that the city and country names compared in the first four task categories only have a single sense (referring to the respective location), while the words asked for in other task categories can have multiple meanings. For example, ``run'' can be used as both a noun or a verb, additionally, in some contexts it refers to the sport activity while other times it is used in a more abstract sense, e.g.~in the context of someone running for president. Therefore, the results in the other task categories might improve if the words' context vectors are first clustered and then the ConEc embedding is generated by multiplying the word2vec embeddings with the average of only those context vectors corresponding to the word's sense most appropriate for the task category.

\begin{table*}[!t]
  \caption{Accuracy on the analogy task with mean and standard deviation computed using three random seeds when initializing the word2vec model. The best results for each category and corpus are in bold.}
  \label{analogy}
  \centering\small
  \begin{tabular}{lr@{$\pm$}lr@{$\pm$}lr@{$\pm$}lr@{$\pm$}l}

    \toprule
   & \multicolumn{4}{c}{text8 (10 iter)}&\multicolumn{4}{c}{1-billion}\\
    \cmidrule(lr){2-5}\cmidrule(lr){6-9}
                    & \multicolumn{2}{c}{word2vec} & \multicolumn{2}{c}{Context Encoder} & \multicolumn{2}{c}{word2vec} & \multicolumn{2}{c}{Context Encoder}\\
    \midrule
capital-common-countries    & 63.8 & 4.7 & $\quad\;$\textbf{78.7} & 0.2 & 79.3 & 2.2 & $\quad\;$\textbf{83.1} & 1.2 \\
capital-world               & 34.0 & 2.1 & \textbf{54.7} & 1.3          & 63.8 & 1.4 & \textbf{75.9} & 0.4  \\
currency                    & 15.4 & 0.9 & \textbf{19.3} & 0.6          & 13.3 & 3.6 & \textbf{14.8} & 0.8  \\
city-in-state               & 28.6 & 1.0 & \textbf{43.6} & 0.9          & 19.6 & 1.7 & \textbf{29.6} & 1.0  \\
family                      & \textbf{79.6} & 1.5 & 77.2 & 0.4          & 78.7 & 2.2 & \textbf{79.0} & 1.4  \\
gram1-adjective-to-adverb   & 11.0 & 0.9 & \textbf{16.6} & 0.7          & 12.3 & 0.5 & \textbf{13.3} & 1.1  \\
gram2-opposite              & 24.3 & 3.0 & 24.3 & 2.0                   & \textbf{27.6} & 0.1 & 21.3 & 1.1  \\
gram3-comparative           & \textbf{64.3} & 0.5 & 63.0 & 1.1          & \textbf{83.7} & 0.9 & 76.2 & 1.1  \\
gram4-superlative           & \textbf{40.3} & 2.1 & 37.6 & 1.5          & \textbf{69.4} & 0.5 & 56.2 & 1.2  \\
gram5-present-participle    & 30.5 & 1.0 & \textbf{31.7} & 0.4          & \textbf{78.4} & 1.0 & 68.0 & 0.7  \\
gram6-nationality-adjective & \textbf{70.6} & 1.5 & 67.2 & 1.4          & 83.8 & 0.6 & \textbf{83.8} & 0.5  \\
gram7-past-tense            & 30.5 & 1.8 & \textbf{33.0} & 0.6          & \textbf{53.9} & 0.9 & 49.2 & 0.7  \\
gram8-plural                & \textbf{49.8} & 0.3 & 49.2 & 1.2          & \textbf{62.7} & 1.9 & 56.7 & 1.0  \\
gram9-plural-verbs          & \textbf{41.0} & 2.5 & 30.1 & 1.9          & \textbf{68.7} & 0.2 & 45.0 & 0.4  \\
\cmidrule{1-9}
total                       & 42.1 & 0.6 & \textbf{46.5} & 0.1          & \textbf{57.2} & 0.3 & 55.8 & 0.3  \\
    \bottomrule
  \end{tabular}
\end{table*}

\end{document}